\documentclass{article}

\usepackage{times}
\usepackage[accepted]{icml2016}

\usepackage{graphicx} 
\usepackage{subfigure}

\usepackage{natbib}

\usepackage{algorithm}
\usepackage{algorithmic}

\usepackage{hyperref}

\usepackage{icml2016}

\usepackage{times}
\usepackage{epsfig}
\usepackage{amsmath}
\usepackage{amssymb}
\usepackage{color}
\usepackage{perpage} 
\MakeSorted{figure} 

\icmltitlerunning{LLNet: Deep Autoencoders for Low-light Image Enhancement}

\begin{document}

\twocolumn[
\icmltitle{LLNet: A Deep Autoencoder approach to Natural Low-light Image Enhancement}

\icmlauthor{Kin Gwn Lore}{kglore@iastate.edu}
\icmladdress{Iowa State University, Ames IA-50014}
\icmlauthor{Adedotun Akintayo}{akintayo@iastate.edu}
\icmladdress{Iowa State University, Ames IA-50014}
\icmlauthor{Soumik Sarkar}{soumiks@iastate.edu}
\icmladdress{Iowa State University, Ames IA-50014}

\icmlkeywords{boring formatting information, machine learning, ICML}

\vskip 0.3in
]

\begin{abstract}
This paper proposes a deep autoencoder-based approach to identify signal features from low-light images and adaptively brighten images without over-amplifying/saturating the lighter parts in images with a high dynamic range. In surveillance, monitoring and tactical reconnaissance, gathering visual information from a dynamic environment and accurately processing such data are essential to making informed decisions and ensuring the success of a mission. Camera sensors are often cost-limited to capture clear images or videos taken in a poorly-lit environment. Many applications aim to enhance brightness, contrast and reduce noise content from the images in an on-board real-time manner. We show that a variant of the stacked-sparse denoising autoencoder can learn to adaptively enhance and denoise from synthetically darkened and noise-added training examples. The model can be applied to images taken from natural low-light environment and/or are hardware-degraded. Results show significant credibility of the approach both visually and by quantitative comparison with various image enhancement techniques.
\end{abstract}

\section{Introduction and motivation}
Good quality images and videos are key to critical automated and human-level decision-making for tasks ranging from security applications, military missions, path planning to medical diagnostics and commercial recommender systems. Clean, high-definition pictures captured by camera systems provide better evidence for well-informed course of action. However, cost constraints limit large scale applications of such systems, thus inexpensive sensors are usually employed. Insufficient lighting, in addition to low sensor quality, produce image noise that may inhibit Intelligence, Surveillance \& Reconnaissance (ISR) missions and commercial industries. These applications stand to benefit from improvements in image enhancement algorithms.


Recently, deep learning-based approaches gained immense traction as they are shown to outperform other state-of-the-art machine learning tools in many computer vision applications, including object recognition~\cite{KSH12}, scene understanding~\cite{CFN13} and occlusion detection~\cite{SVR15}. While neural networks have been widely studied for image denoising tasks, there are no existing works using deep networks to both enhance and denoise images taken in poorly illuminated environments. We approach the problem of low-light image enhancement from a representation learning perspective using deep autoencoders (we refer to as Low-light Net, LLNet) trained to learn underlying signal features in low-light images and adaptively brighten and denoise. The method takes advantage of local path-wise contrast improvement similar to the works in~\cite{LBHA13} such that improvements are done relative to local neighbors to prevent over-amplifying already-bright pixels. The same network is trained to learn noise structures in order to produce a brighter, denoised image.


\textbf{Contributions:}
The present paper presents a novel application of using a class of deep neural networks--stacked sparse denoising autoencoder (SSDA)--to enhance natural low-light images. We propose a training data generation method by synthetically modifying images available on Internet databases to simulate low-light environments. Two types of deep architecture are explored - (i) for simultaneous learning of contrast-enhancement and denoising (LLNet) and (ii) sequential learning of contrast-enhancement and denoising using two modules (\textit{staged} LLNet or S-LLNet). The performance of the trained networks are evaluated and compared against other methods on test data with synthetic noise and artificial darkening along with. The same procedure is repeated on natural low-light images to demonstrate the enhancement capability of the synthetically trained model applied on a realistic set of images obtained with regular cell-phone camera in low-light environments. Hidden layer weights of the deep networks are visualized to offer insights to the features learned by the models.

\section{Related work}

There are well-known contrast enhancement methods such as improving image contrast by histogram equalization~\cite{trahanias1992color,cheng2004simple,pizer1987adaptive}. Contrast-limiting Adaptive Histogram Equalization (CLAHE)~\cite{PZHDJMBP98} belongs to the class of histogram-stretching methods and serves to limit to the extent of contrast enhancement result of histogram equalization. Subsequently, an optimization technique, OCTM~\cite{XW11} was introduced for mapping the contrast-tone of an image with the use of mathematical transfer function. However, this requires weighting of some domain knowledge as well as an associated complexity increase. Available schemes also explored using non-linear functions like the gamma function~\cite{GW01} to enhance image contrast. Histogram stretching methods~\cite{KT11} and its variants like Brightness preserving Bi-histogram Equalization (BBHE) and Quantized Bi-histogram Equalization (QBHE)~\cite{KKK11} gained prominence to improve on the artifacts of histogram equalization. Also,


Image denoising tasks have been explored using BM3D~\cite{dabov2009bm3d,dabov2008image,dabov2007image}, K-SVD~\cite{elad2006image}, and non-linear filters~\cite{chen1999tri,chan2005salt}. Using deep learning, authors in~\cite{VLB08} presented the concept of denoising autoencoders for learning features from noisy images while ~\cite{JS08} applied convolutional neural networks to denoise natural images. The network was applied for inpainting~\cite{xie2012image} and deblurring~\cite{SHHS14}. In addition, authors in~\cite{agostinelli2013adaptive} implemented an adaptive multi-column architecture to robustly denoise images by training the model with various types of noise and testing on images with arbitrary noise levels and types. Stacked denoising autoencoders were used in~\cite{burger2012image} to reconstruct clean images from noisy images by exploiting the encoding layer of the multilayer perceptron (MLP).

\section{The Low-light Net (LLNet)}
The proposed framework is introduced in this section along with training methodology and network parameters.

\textbf{Learning features from low-light images:}
SSDAs are sparsity-inducing variant of deep autoencoders that ensures learning the invariant features embedded in the proper dimensional space of the dataset in an unsupervised manner. Early proponents~\cite{VLB08} have shown that by stacking several denoising autoencoders (DA) in a greedy layer-wise manner for pre-training, the network is able to find a better parameter space during error back-propagation.

Let $\textbf{y}\in \mathcal{R}^N$ be the clean, uncorrupted data and $\textbf{x}\in \mathcal{R}^N$ be the corrupted, noisy version of $\textbf{y}$ such that $\textbf{x} = \textbf{My}$, where $\textbf{M}\in \mathcal{R}^{N \times N}$ is the high-dimensional, non-analytic matrix assumed to have corrupted the clean data. With DA, feed-forward learning functions are defined to characterize each element of $\textbf{M}$ by the expression $h(\textbf{x}) = \sigma(\textbf{Wx}+\textbf{b})$ and $\hat{\textbf{y}}(\textbf{x}) = \sigma'(\textbf{W}'\textbf{h}+\textbf{b}')$, where $\sigma$ and $\sigma'$ denote the encoding and decoding functions (either of which is usually the sigmoid function $\sigma(\textbf{s})$ or $\sigma'(\textbf{s})= (1+\exp(-\textbf{s}))^{-1}$ of a single DA layer with $K$ units, respectively. $\textbf{W} \in \mathcal{R}^{K\times N}$ and $\textbf{b} \in \mathcal{R}^K$ are the weights and biases of each layers of encoder whereas $\textbf{W}' \in \mathcal{R}^{N\times K}$ and $\textbf{b} \in \mathcal{R}^K$ are the weights and biases for each layer of the decoder. $h(\textbf{x})\in \mathcal{R}^K$ is the activation of the hidden layer and $\hat{\textbf{y}}(\textbf{x})\in \mathcal{R}^N$ is the reconstruction of the input (\textit{i.e.} the output of the DA).

\begin{figure*}[lhtb]
\centering
 \includegraphics[width=0.6\textwidth,trim={0 0 50 0}]{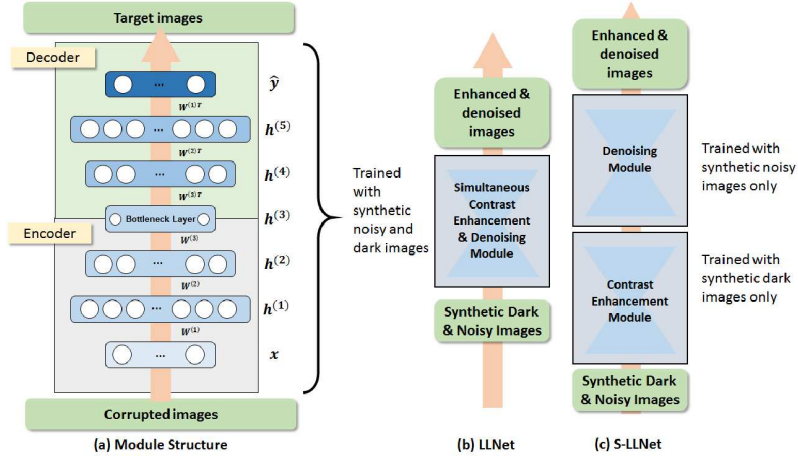}
\caption{Architecture of the proposed framework: (a) An autoencoder module is comprised of multiple layers of hidden units, where the encoder is trained by unsupervised learning, the decoder weights are transposed from the encoder and subsequently fine-tuned by error backpropagation; (b) LLNet with a simultaneous contrast-enhancement and denoising module; (c) S-LLNet with sequential contrast-enhancement and denoising modules. The purpose of denoising is to remove noise artifacts often accompanying contrast enhancement.}
\label{fig:ssda}
\end{figure*}

%

LLNet framework takes its inspiration from SSDA whose sparsity-inducing characteristic aids learning features to denoise signals. We take the advantage of SSDA's denoising capability and the deep network's complex modeling capacity to learn features underlying in low-light images and produce enhanced images with minimal noise and improved contrast. The key aspect is that the network is trained using images obtained from internet databases that are subsequently synthetically processed (i.e. darkening nonlinearly and adding Gaussian noise) to simulate low-light condition, since collection of a large number of natural low-light images (sufficient for deep network training) and their well-lit counterparts can be impractical. In our experiments, both synthetic and natural images are used to evaluate the network's performance in denoising and contrast-enhancement.

Aside from the regular LLNet where the network is trained with both darkened and noisy images, we also propose the \textit{staged} LLNet (S-LLNet) containing separate modules arranged in series for contrast enhancement (stage 1) and denoising (stage 2). The key distinction over the regular LLNet is that the modules are trained separately with darkened-only training sets and noisy-only training sets. Both structures are presented in Fig.~\ref{fig:ssda}. While the S-LLNet architecture provides a greater flexibility of training (and certain performance improvement as shown in Section~\ref{sec:res}), it increases the inference time slightly which may be a concern for certain real-time applications. However, customized hardware-acceleration can resolve such issues significantly.

\textbf{Network parameters: }
LLNet is comprised of 3 DA layers, with the first DA layer taking the input image of dimensions $17\times17$ pixels (i.e. $289$ input units). The first DA layer has 867 hidden units, the second has 578 hidden units, and the third has 289 hidden units which becomes the bottleneck layer. Beyond the third DA layer forms the decoding counterparts of the first three layers, thus having 578 and 867 hidden units for the fourth and fifth layers respectively. Output units have the same dimension as the input, i.e. 289. The network is pre-trained for 30 epochs with pre-training learning rates of 0.1 for the first two DA layers and 0.01 for the last DA layer, whereas finetuning was performed with a learning rate of 0.1 for the first 200 finetuning epochs, 0.01 afterwards, and stops only if the improvement in validation error is less than 0.5\%. For the case of S-LLNet, the parameters of each module are identical.

\textbf{Training the model: }
Training was performed using 422,500 patches, extracted from 169 standard test images\footnote {Dataset URL: http://decsai.ugr.es/cvg/dbimagenes/}. Consistent with current practices, the only pre-processing done was to normalize the image pixels to between zero and one. During the generation of the patches, we produced 2500 patches from random locations (and with random darkening and noise parameters) from the same image. The $17\times17$ pixel patches are then darkened nonlinearly using the MATLAB command \texttt{imadjust} to randomly apply a gamma adjustment. Gamma correction is a simple but general case with application of a power law formula to images for pixel-wise enhancement with $I_\text{out} = A \times I_\text{in}^{\gamma}$, where $A$ is a constant determined by the maximum pixel intensity in the image. Intuitively, image is brightened when $\gamma<1$ while $\gamma=1$ leaves it unaffected. Therefore, when $\gamma>1$, the mapping is weighted toward lower (darker) grayscale pixel intensity values.

A uniform distribution of $\gamma\sim\text{Uniform}~(2,5)$ with random variable $\gamma$, is selected to result in training patches that are darkened to a varying degree, thus simulating multiple low-light scenarios possible in real-life. Additionally, to simulate low quality cameras used to capture images, these training patches are further corrupted by a Gaussian noise via MATLAB function \texttt{imnoise} with standard deviation of $\sigma=\left(B(25/255)^2\right)^{1/2}$, where $B\sim\text{Uniform}~(0,1)$. Hence, the final corrupted image and the original image exhibit the following relationship taking the form of $I_\text{train} = n\left(g(I_\text{original})\right)$ where function $g(\cdot)$ represents the gamma adjustment function and $n(\cdot)$ represents the noise function.

Random gamma darkening with random noise levels result in a variety of training images that can increase the robustness of the model. In reality, natural low-light images may also include quantization and Poisson noise in addition to Gaussian noise. We chose a Gaussian-only model for the ease of analysis and as a preliminary feasibility study of the framework trained on synthetic images and applied to natural images. Furthermore, since Gaussian noise is a very familiar yet popular noise model for many image denoising tasks, we can acquire a sense of how well LLNet performs with respect to other image enhancement algorithms. The patches are randomly shuffled and then divided into 211,250 training examples and 211,250 validation samples. Training involves learning the invariant representation of low light and noise with the autoencoder to eventually denoise and simultaneously enhance the contrast of these darkened patches. The reconstructed image is compared against the clean version (i.e. bright, noiseless image) by computing the mean-squared error.


When training both LLNet and S-LLNet, each DA is trained by error back-propagation to minimize the sparsity regularized reconstruction loss as described in Xie et al.~\cite{xie2012image}, expressed as $
\mathcal{L}_{\text{DA}}(\mathcal{D};\theta) = \frac{1}{N} \sum_{i=1}^N \frac{1}{2} || y_i - \hat{y}(x_i) ||_2^2 + \beta \sum_{j=1}^K \text{KL}(\hat{\rho}_j||\rho) + \frac{\lambda}{2}(||W||_F^2 + ||W'||_F^2)
$ where $N$ is the number of patches, $\theta=\{W,b,W',b'\}$ are the parameters of the model, KL$(\hat{\rho}_j || \rho) $ is the Kullback-Leibler divergence between $\rho$ (target activation) and $\hat{\rho}_j$ (empirical average activation of the $j$-th hidden unit) which induces sparsity in the hidden layers with $\text{KL}(\hat{\rho}_j || \rho) = \rho \log \frac{\rho}{\hat{\rho}_j}+(1-\rho)\log{\frac{1-\rho}{1-\hat{\rho}_j}}$ where $\hat{\rho}_j = \frac{1}{N} \sum_{i=1}^N h_j (x_i)$. $\lambda, \beta$ and $\rho$ are scalar hyper-parameters determined by cross-validation.

After the weights of the decoder have been initialized, the entire pretrained network is finetuned using an error back-propagation algorithm to minimize the loss function given by $\mathcal{L}_{\text{SSDA}}(\mathcal{D};\theta) = \frac{1}{N} \sum_{i=1}^N || y_i - \hat{y}(x_i) ||_2^2 + \frac{\lambda}{L}\sum_{l=1}^{2L}||W^{(l)}||_F^2 $ where $L$ is the number of stacked DAs and $W^{(l)}$ denotes weights for the $l$-th layer in the stacked deep network. The sparsity inducing term is not needed for this step because the sparsity was already incorporated in the pre-trained DAs.

\textbf{Image reconstruction:} During inference, the test image is first broken up into overlapping $17\times 17$ patches with stride size of $3\times3$. The collection of patches is passed through LLNet to obtain corresponding denoised, contrast-enhanced patches. The patches are re-arranged back into its original dimensions where overlapping regions are averaged. We find that using a patching stride of $2\times2$ or even $1\times1$ (fully overlapped patches) do not produce significantly superior results. Additionally, increasing the number of DA layers improves the nonlinear modeling capacity of the network. However, training a larger model is computationally more expensive and we determined that the current network structure is adequate for the present study.

\section{Evaluation metrics and compared methods}
In this section we present brief descriptions of other contrast-enhancement methods along with the performance metric used to evaluate the proposed framework's performance.

\begin{figure}[lhtb]
\centering
 \includegraphics[width=0.5\textwidth,trim={0 60 0 30}]{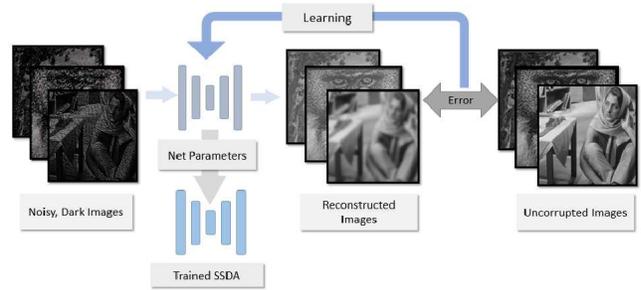}
\caption{Training the LLNet: Training images are synthetically darkened and added with noise. These images are fed through LLNet where the reconstructed images are compared with the uncorrupted images to compute the error, which is then backpropagated to finetune and optimize the model weights and biases.}
\label{fig:ssda_train}
\end{figure}

\subsection{Performance metric}
Two metrics are used, namely the peak signal-to-noise ratio (PSNR) and the structural similarity index (SSIM).

\textbf{Peak signal-to-noise ratio (PSNR):}
PSNR quantifies the extent of corruption of original image with noise as well as approximating human perception of the image. It has also been established to demonstrate direct relationship with compression-introduced noise~\cite{SNSH11}. Roughly, the higher the PSNR, the better the denoised image especially with the same compression code.

\textbf{Structural similarity index (SSIM):}
SSIM a metric for capturing the perceived quality of digital images and videos~\cite{LBHA13,WBSS04}. It is used to measure the similarity between two images. SSIM quantifies the measurement or prediction of image quality with respect to initial uncompressed or distortion-free image as reference. Although PSNR and MSE are known to quantify the absolute error between the result and the reference image, such metrics may not really quantify complete similarity. On the other hand, SSIM explores the change in image structure and being a perception-type model, it incorporates pixel inter-dependencies as well as masking of contrast and pixel intensities.

\subsection{Compared methods}
This subsection describes certain popular methods for enhancing low-light images used here for comparison.

\textbf{Histogram equalization (HE):} Histogram equalization usually increases the global contrast of images, especially when the usable data of the image is represented by close contrast values. Through this adjustment, the intensities can be better distributed on the histogram. This allows for areas of lower local contrast to gain a higher contrast. Histogram equalization accomplishes this by effectively spreading out the most frequent intensity values. The method is useful in images with backgrounds and foregrounds that are both bright or both dark. 

\textbf{Contrast-limiting adaptive histogram equalization (CLAHE):} Contrast-limiting adaptive histogram equalization differs from ordinary adaptive histogram equalization in its contrast limiting. In the case of CLAHE, the contrast limiting procedure has to be applied for each neighborhood from which a transformation function is derived. CLAHE was developed to prevent the over-amplification of noise that arise in adaptive histogram equalization.

\textbf{Gamma adjustment (GA):} Gamma curves illustrated with $\gamma>1$ have exactly the opposite effect as those generated with $\gamma<1$. It is important to note that gamma correction reduces toward the identity curve when $\gamma=1$. As discussed earlier, the image is brightened when $\gamma<1$, darkened with $\gamma>1$, while $\gamma=1$ leaves it unaffected.

\textbf{Histogram equalization with 3D block matching (HE+BM3D):} BM3D is the current state-of-the-art algorithm for image noise removal presented by ~\cite{dabov2007image}. It uses a collaborative form of Wiener filter for high dimensional block of patches by grouping similar 2D blocks into a 3D data array. The algorithm ensures the sparsity in transformed domain and takes advantage of joint denoising of grouped patches similar in ways to pixel-wise overcompleteness which K-Singular value Decomposition (KSVD)~\cite{elad2006image} (the former best performing denoising method), ensured on patch-based dictionaries. Finally, domain inversion is done and the results of different matched block are fused together. In this work we attempt to first equalize the contrast of the test image, then use BM3D as a denoiser to remove the noise resulting from histogram equalization. We attempted to reverse the order, i.e. use BM3D to remove noise from the low-light images first then apply contrast enhancement. As BM3D removes noise by patching images, the patch boundaries get significantly amplified and become extremely pronounced when histogram equalization is applied, hence, producing non-competitive results.

\section{Results and discussion}\label{sec:res}
\begin{figure}[!t]
\centering
 \includegraphics[width=0.47\textwidth,trim={0 35 0 0}]{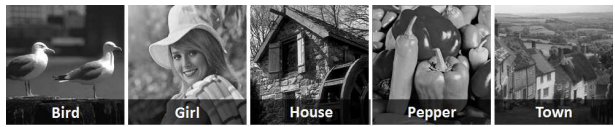}
\caption{Original standard test images used to compute PSNR and SSIM.}
\vspace{-20pt}
\label{fig:originalimages}
\end{figure}

In this section, we evaluate the performance of our framework against the methods outlined above on standard images shown in Fig.~\ref{fig:originalimages}. Test images are darkened with $\gamma=3$, where noisy versions contain Gaussian noise of $\sigma=18$ and $\sigma=25$, which are typical values for image noise under poor illumination and/or high temperature; these parameters correspond to scaled variances of $\sigma_s^2=0.005$ and $\sigma_s^2=0.010$ respectively if the pixel intensities are in 8-bit integers ($\sigma_s = \sigma/255$ where $\sigma_s\in[0,1]$ and $\sigma\in[0,255]$). Histogram equalization is performed by using the MATLAB function \texttt{histeq}, whereas CLAHE is performed with the function \texttt{adapthisteq} with default parameters ($8\times8$ image tiles, contrast enhancement limit of 0.01, full range output, 256 bins for building contrast enhancing transformation, uniform histogram distribution, and distribution parameter of 0.4). Gamma adjustment is performed on the test image with $\gamma=0.3$. For the hybrid `HE+BM3D' method, we first applied histogram equalization to enhance image contrast before using the BM3D code developed by Dabov et al.~\cite{dabov2007image} as a denoiser, where the noise standard deviation input parameter for BM3D is set to $\sigma=25$ (the highest noise level of the test image). Both LLNet and S-LLNet outputs are reconstructed with overlapping $17\times 17$ patches of stride size $3\times 3$. Training was performed on NVIDIA's TITAN X GPU using Theano's deep learning framework~\cite{Bastien-Theano-2012,bergstra+al:2010-scipy} and took approximately 30 hours. Enhancing an image with dimension of $512\times512$ pixels took 0.42 $s$ on GPU.

\begin{figure*}[!htb]
\centering
 \includegraphics[width=0.8\textwidth,trim={0 0 0 0}]{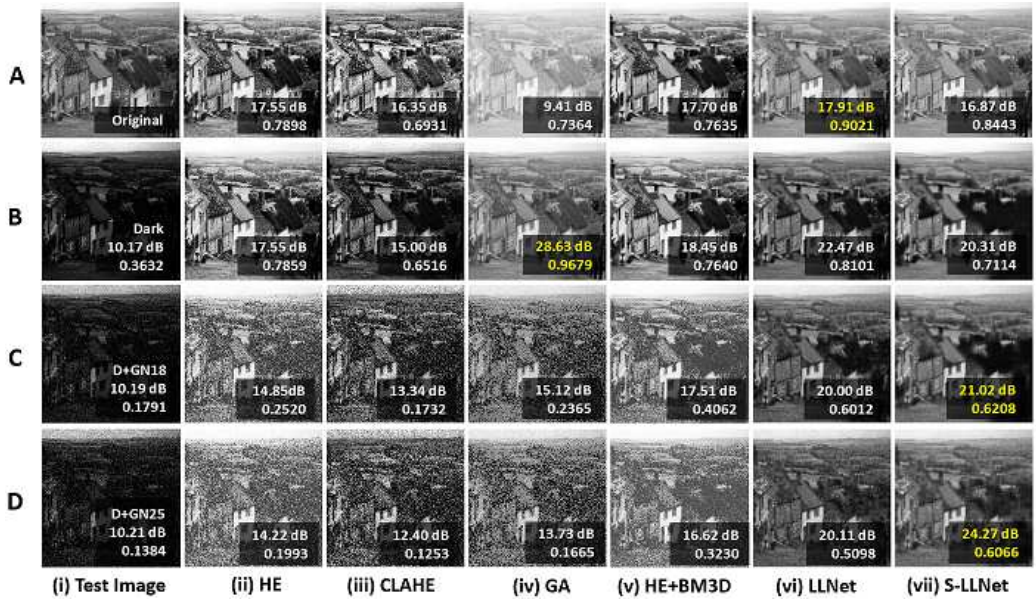}
\caption{Comparison of methods of enhancing `Town' when applied to (A) original already-bright, (B) darkened, (C) darkened and noisy $(\sigma=18)$, and (D) darkened and noisy $(\sigma=25)$ images. Darkening is done with $\gamma=3$. The numbers with units dB are PSNR, the numbers without are SSIM. Best viewed on screen.}
\label{fig:llnet_synthetic}
\end{figure*}


\begin{table*}
\caption{PSNR and SSIM of outputs using different enhancement methods. `Bird' means the non-dark and noiseless (i.e. original) image of Bird. `Bird-D' indicates a darkened version of the same image. `Bird-D+GN18' denotes a darkened Bird image with added Gaussian noise of $\sigma=18$, whereas `Bird-D+GN25' denotes darkened Bird image with added Gaussian noise of $\sigma=25$. Bolded numbers correspond to the method with the highest PSNR or SSIM. Asterisk (*) denotes our framework.}
\begin{center}
{\scriptsize
\begin{tabular}{|l|c|c|c|c|c|c|c|}\hline
\textbf{PSNR (dB)/SSIM} & \textbf{Test} & \textbf{HE} & \textbf{CLAHE} & \textbf{GA} & \textbf{HE+BM3D}& \textbf{LLNet*} & \textbf{S-LLNet*}\\\hline
\textbf{Bird}&	N/A	/	1.00	&	11.22	/	0.63	&	\textbf{21.55}	/	0.90	&	8.26	/	0.63	&	11.27	/	0.69	&	20.35	/	\textbf{0.92}	&	18.07	/	0.87\\
\textbf{Bird-D}&	12.27	/	0.18	&	11.28	/	0.62	&	15.15	/	0.52	&	\textbf{26.06}	/	\textbf{0.84}	&	11.35	/	0.71	&	18.43	/	0.60	&	16.18	/	0.53\\
\textbf{Bird-D+GN18}&	12.56	/	0.14	&	9.25	/	0.09	&	14.63	/	0.11	&	13.49	/	0.10	&	9.98	/	0.13	&	\textbf{19.73}	/	\textbf{0.56}	&	18.60	/	0.54\\
\textbf{Bird-D+GN25}&	12.70	/	0.12	&	9.04	/	0.08	&	13.60	/	0.09	&	12.51	/	0.08	&	9.72	/	0.11	&	21.17	/	0.50	&	\textbf{22.11}	/	0.61\\\hline

\textbf{Girl}&	N/A	/	1.00	&	\textbf{18.24}	/	0.80	&	17.02	/	0.70	&	10.52	/	0.78	&	18.23	/	0.69	&	17.33	/	\textbf{0.84}	&	14.57	/	0.76\\
\textbf{Girl-D}&	9.50	/	0.50	&	18.27	/	0.80	&	14.36	/	0.66	&	\textbf{29.73}	/	\textbf{0.99}	&	18.26	/	0.69	&	22.45	/	0.80	&	21.62	/	0.74\\
\textbf{Girl-D+GN18}&	9.43	/	0.21	&	16.07	/	0.26	&	12.95	/	0.17	&	16.90	/	0.30	&	19.28	/	0.53	&	20.04	/	0.60	&	\textbf{22.07}	/	\textbf{0.66}\\
\textbf{Girl-D+GN25}&	9.39	/	0.15	&	15.33	/	0.19	&	12.09	/	0.12	&	15.14	/	0.20	&	18.50	/	0.39	&	19.60	/	0.49	&	\textbf{22.68}	/	\textbf{0.60}\\\hline

\textbf{House}&	N/A	/	1.00	&	13.36	/	0.70	&	\textbf{18.89}	/	\textbf{0.81}	&	9.50	/	0.56	&	13.24	/	0.61	&	11.61	/	0.60	&	10.57	/	0.50\\
\textbf{House-D}&	12.12	/	0.33	&	12.03	/	0.65	&	16.81	/	0.60	&	\textbf{26.79}	/	\textbf{0.82}	&	11.92	/	0.54	&	21.10	/	0.64	&	18.73	/	0.49\\
\textbf{House-D+GN18}&	12.19	/	0.29	&	10.55	/	0.33	&	15.48	/	0.35	&	13.76	/	0.33	&	11.39	/	0.42	&	\textbf{20.25}	/	\textbf{0.56}	&	19.91	/	0.51\\
\textbf{House-D+GN25}&	12.16	/	0.26	&	10.09	/	0.29	&	14.08	/	0.29	&	12.67	/	0.28	&	10.94	/	0.37	&	19.71	/	\textbf{0.52}	&	\textbf{20.76}	/	0.51\\\hline

\textbf{Pepper}&	N/A	/	1.00	&	\textbf{18.61}	/	\textbf{0.90}	&	18.27	/	0.76	&	9.63	/	0.69	&	18.61	/	0.84	&	10.92	/	0.71	&	9.93	/	0.66\\
\textbf{Pepper-D}&	10.45	/	0.37	&	18.45	/	0.85	&	15.46	/	0.61	&	\textbf{28.28}	/	\textbf{0.90}	&	18.45	/	0.80	&	21.33	/	0.78	&	19.54	/	0.72\\
\textbf{Pepper-D+GN18}&	11.13	/	0.19	&	14.69	/	0.21	&	14.47	/	0.17	&	14.66	/	0.22	&	16.97	/	0.57	&	\textbf{22.23}	/	0.57	&	21.80	/	\textbf{0.65}\\
\textbf{Pepper-D+GN25}&	10.41	/	0.15	&	13.67	/	0.15	&	13.31	/	0.13	&	13.87	/	0.16	&	15.96	/	0.36	&	21.48	/	0.48	&	\textbf{23.38}	/	\textbf{0.62}\\\hline

\textbf{Town}&	N/A	/	1.00	&	17.55	/	0.79	&	16.35	/	0.69	&	9.41	/	0.74	&	17.70	/	0.76	&	\textbf{17.91}	/	\textbf{0.90}	&	16.87	/	0.84\\
\textbf{Town-D}&	10.17	/	0.36	&	17.55	/	0.79	&	15.00	/	0.65	&	\textbf{28.63}	/	\textbf{0.97}	&	17.72	/	0.76	&	22.47	/	0.81	&	20.31	/	0.71\\
\textbf{Town-D+GN18}&	10.19	/	0.18	&	14.85	/	0.25	&	13.34	/	0.17	&	15.12	/	0.24	&	17.51	/	0.41	&	20.00	/	0.60	&	\textbf{21.02}	/	\textbf{0.62}\\
\textbf{Town-D+GN25}&	10.21	/	0.14	&	14.22	/	0.20	&	12.40	/	0.13	&	13.73	/	0.17	&	16.62	/	0.32	&	20.11	/	0.51	&	\textbf{24.27}	/	\textbf{0.61}

 \\\hline
\end{tabular}
}
\end{center}
\vspace{-15pt}
\label{tab:psnr_ssim_images}
\end{table*}

\begin{figure*}[!htb]
\centering
 \includegraphics[width=0.8\textwidth,trim={0 0 0 0}]{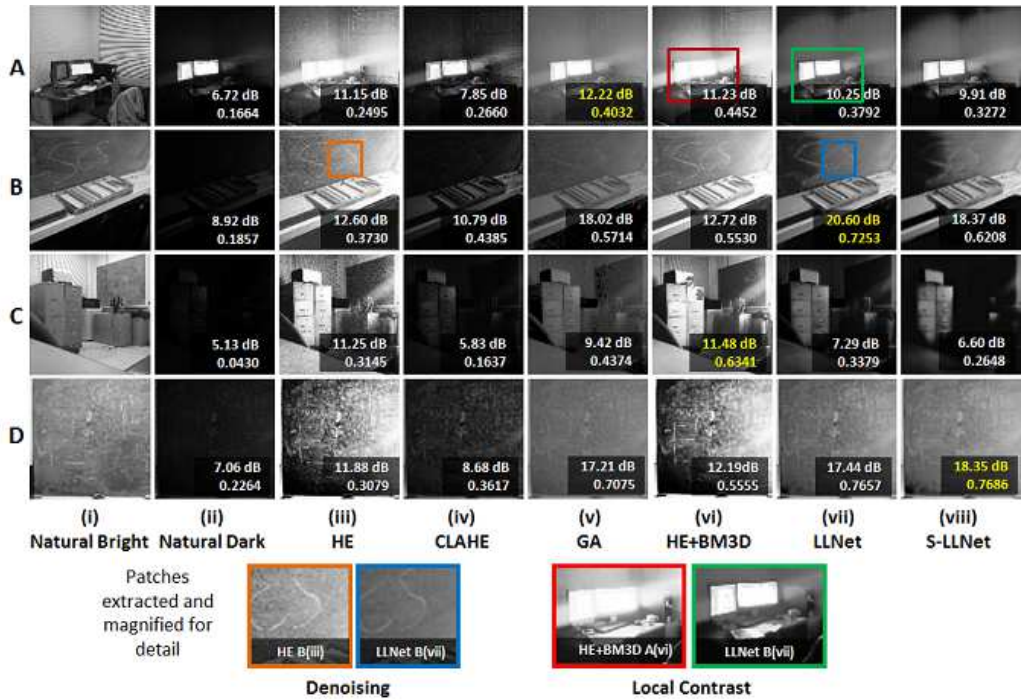}
\caption{Comparison of methods of enhancing naturally dark images of (A) computer, (B) chalkboard, (C) cabinet, and (D) writings. Selected regions are enlarged to demonstrate the denoising and local contrast enhancement capabilities of LLNet. HE (including HE+BM3D) results in overamplification of the light from the computer display whereas LLNet was able to avoid this issue. In the natural case, the best performers in terms of PSNR and SSIM coincide. Best viewed on screen.}
\vspace{-10pt}
\label{fig:llnet_darknatural}
\end{figure*}

\textbf{Algorithm adaptivity: }
Ideally, an already-bright image should no longer be brightened any further. To test this, different enhancement algorithms are performed on a normal, non-dark and noiseless image. Fig.~\ref{fig:llnet_synthetic}A shows the result when running the `Town' image through various algorithms. LLNet outputs a slightly brighter image, but not to the degree that everything appears over-brightened and washed-out like that in GA output. This shows that in the process of learning low-light features, LLNet successfully learns the necessary degree of required brightening that should be applied to the image. However, when evaluating contrast enhancement via visual inspection, histogram equalization methods (i.e. HE, CLAHE, HE+BM3D) provide superior enhancement given the original image. When tested with other images (namely, `Bird', `Girl', `House', `Pepper', etc.) as shown in Table~\ref{tab:psnr_ssim_images}, HE-based methods generally fared slightly better with higher PSNR and SSIM.

\textbf{Enhancing artificially darkened images: }
Fig.~\ref{fig:llnet_synthetic}B shows output of various methods when enhancement is applied to a `Town' image darkened with $\gamma=3$. Here, LLNet achieves the highest PSNR followed by GA, but the the other way round when evaluated with SSIM. The similarity of GA-enhanced image with the original is expected because gamma readjustment with $\gamma=0.3\simeq1/3$ essentially reverses the process close to the original intensity levels. In fact, when tested with other images, the highest scores for darkened-only images are achieved only by one of LLNet, S-LLNet or GA. Note that LLNet is trained on varying degrees of $\gamma$ but not with a fixed $\gamma=3$. Results tabulated in Table~\ref{tab:psnr_ssim_images} highlights the advantages and broad applicability of the deep autoencoder approach with LLNet and S-LLNet.

\textbf{Enhancing darkened images in the presence of synthetic noise: }
To simulate dark images taken with regular or subpar camera sensors, Gaussian noise is added to the synthetic dark images. Fig.~\ref{fig:llnet_synthetic}C and~\ref{fig:llnet_synthetic}D presents a gamma-darkened `Town' image corrupted with Gaussian noise of $\sigma=18$ and $\sigma=25$, respectively. For this test image, S-LLNet attains the highest PSNR and SSIM followed by LLNet for both noise levels. Table~\ref{tab:psnr_ssim_images} shows that no other methods aside from LLNet/S-LLNet result in the highest PSNR and SSIM for dark, noisy images. Histogram equalization methods fail due to the intensity of noisy pixels being equalized and produced detrimental effects to the output images. Additionally, BM3D is not able to effectively denoise the equalized images with parameter $\sigma=25$ since the structure of the noise changes during the equalization process.

\textbf{Application on natural low-light images: }
When working with downloaded images, a clean reference image is available for computing PSNR and SSIM. However, reference images may not be available in real life when working with naturally dark images. Since this is a controlled experiment, we circumvented the issue by mounting an ordinary cell-phone (Nexus 4) camera on a tripod to capture pictures in an indoor environment with both lights on and lights off. The picture with lights on are used as the reference images for PSNR and SSIM computations, whereas the picture with lights off becomes the natural low-light test image. Although the bright pictures cannot be considered as the ground truth, it provides a reference point to evaluate the performance of various algorithms. Performance of each enhancement method is shown in Fig.~\ref{fig:llnet_darknatural}. While histogram equalization greatly improves the contrast of the image, it corrupts the output with a large noise content. In addition, the method suffers from over-amplification in regions where there is a very high intensity brightness in dark regions, as shown by blooming effect on the computer display in panel \ref{fig:llnet_darknatural}A(vi) and \ref{fig:llnet_darknatural}A(vii). CLAHE is able to improve the contrast without significant blooming of the display, but like HE it tends to amplify noise within the images. LLNet performs significantly well with its capability to suppress noise in most of the images while improving local contrast, as shown in the magnified patches at the bottom of Fig.~\ref{fig:llnet_darknatural}.

\begin{figure}[!htb]
\centering
 \includegraphics[width=0.5\textwidth,trim={20 50 20 0}]{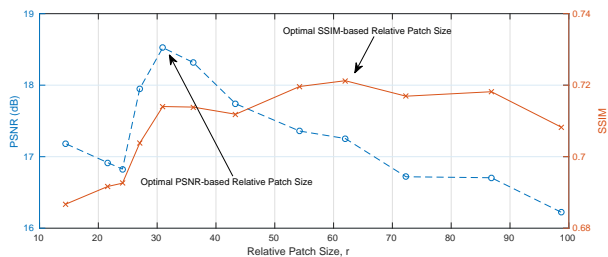}
\caption{Relative patch size vs PSNR and SSIM. The picture with highest PSNR has the highest denoising capability but least sharp. Picture with highest $r$ has the least denoising capability but has the highest image sharpness. Picture with the highest SSIM is more well-balanced.}
\label{fig:dratio}
\end{figure}

\begin{figure*}[!htb]
\centering
 \includegraphics[width=0.8\textwidth,trim={0 10 0 0}]{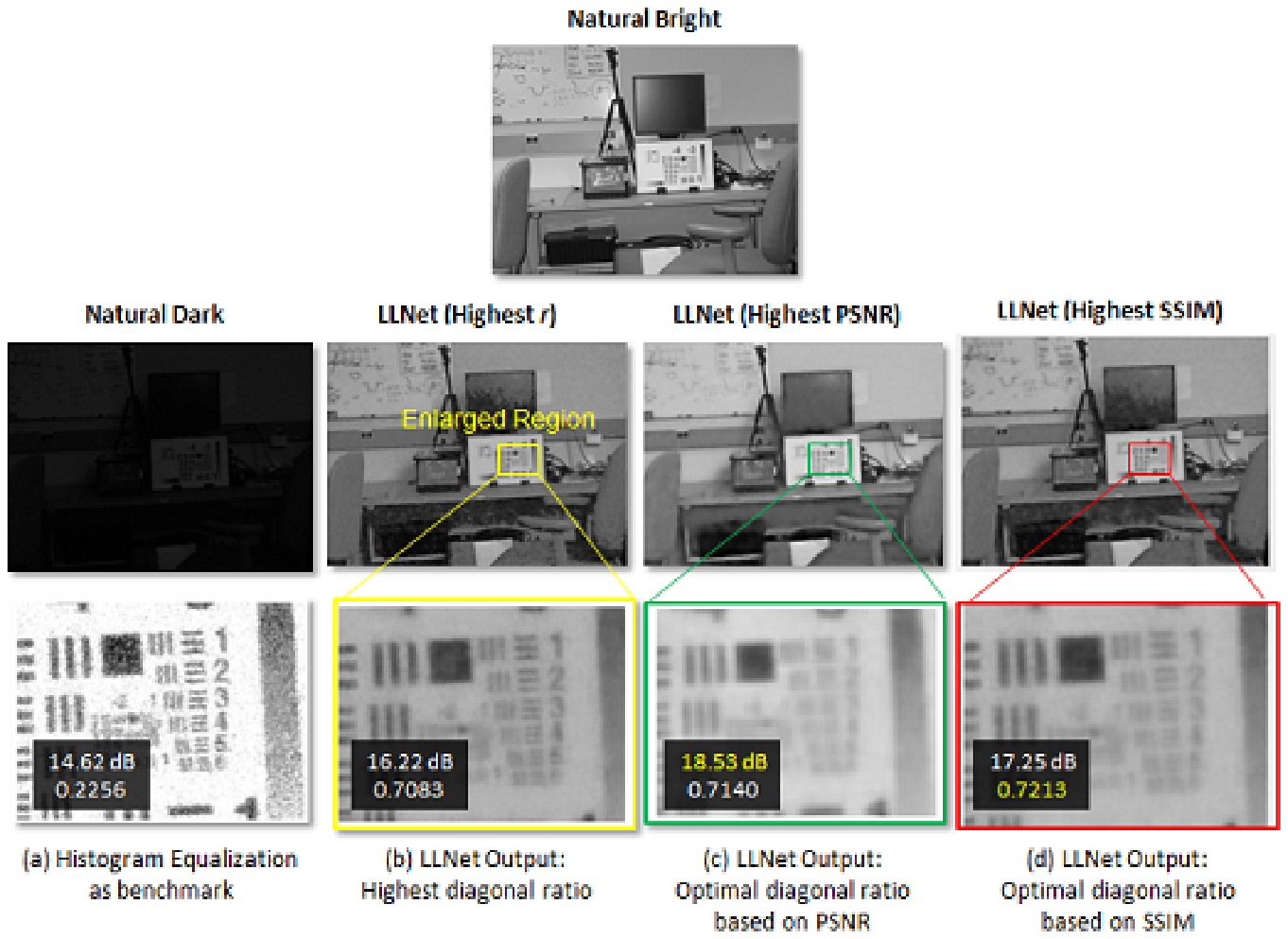}
\caption{Test on natural image with 1951 USAF resolution test chart. There exist optimal relative patch sizes that result in the highest PSNR or SSIM after image enhancement (using LLNet). The result enhanced with histogram equalization is shown to highlight the loss in detail of the natural dark image (where the main light is turned off) compared to the natural bright image.}
\label{fig:labimage}
\end{figure*}

\textbf{Denoising capability, image sharpness, and patch size: }
There is a trade-off between denoising capability and the perceived sharpness of the enhanced image. While higher PSNR indicates a higher denoising capability, this metric favors smoother edges. Therefore, images that are less sharp often achieve a higher PSNR. Hence, SSIM is used as a complementary metric to evaluate the gain or loss in perceived structural information. From the experiments, a relationship between denoising capability (PSNR), similarity levels (SSIM) and image sharpness is found to be dependent on the dimensions of the denoised patch relative to the test image. A smaller patch size implies finer-grain enhancement over the test image, whereas a larger patch size implies coarser enhancement. Because natural images may also come in varying heights and widths, the relative patch size--a dimensionless quantity that relates the patch size to the dimensions of the test image, $r$--is defined as $ r = d_p d_i^{-1} = (w_p^2+h_p^2)^{1/2} (w_i^2 + h_i^2)^{-1/2} $
where quantities $d$, $w$, and $h$ denote the diagonal length, width, and height in pixels, with subscripts $p$ and $i$ referring to the patch and test image respectively. Relative patch size may also be thought as the size of the receptive field on a test image. From the results, it is observed that when the relative patch size decreases, object edges appear sharper at the cost of having more noise. However, there exists an optimal patch size resulting in an enhanced image with the highest PSNR or SSIM (as shown in Fig.~\ref{fig:dratio} and Fig.~\ref{fig:labimage}.). If the optimal patch size is selected based on PSNR, the resulting image will have the lowest noise levels but is less sharp. If the smallest patch size is selected, then the resulting image has the highest sharpness where more details can be observed but with the expense of having more noise. Choosing the optimal patch size based on SSIM produces a more well-balanced result in terms of denoising capability and image sharpness.

We included a natural test image where the US Air Force (USAF) resolution test chart is shown. The test chart consists of groups of three bars varying in sizes labeled with numbers which conforms to the MIL-STD-150A standard set by the US Air Force in 1951. Originally, this test chart is used to determine the resolving power of optical imaging systems such as microscopes, cameras, and image scanners. For the present study, we used this test chart to visually compare the trade-off denoising capability and image sharpness using different relative patch sizes. The results are shown in Fig.~\ref{fig:labimage}.

\textbf{Prior knowledge on input: }
HE can be easily performed on images without any input parameters. Like HE, CLAHE can also be used without any input parameters where the performance can be further finetuned with various other parameters such as tile sizes, contrast output ranges, etc. Gamma adjustment and BM3D both require prior knowledge of the input parameter (values of $\gamma$ and $\sigma$, respectively), thus it is often necessary to finetune the parameters by trial-and-error to achieve the best results. The advantages of using deep learning-based approach, specifically using LLNet and S-LLNet, is that after training them with a large variety of (darkened and noisy) images (with proper choice of hyper-parameters), there is no need for meticulous hand-tuning during testing/practical use. The model automatically extracts and learns the underlying features from low-light images. Essentially, this study shows that a deep model that has been trained with varying degrees of darkening and noise levels can be used for many real-world problems without detail knowledge of camera and environment.

\begin{figure}[!htb]
\centering
 \includegraphics[width=0.5\textwidth,trim={0 40 0 20}]{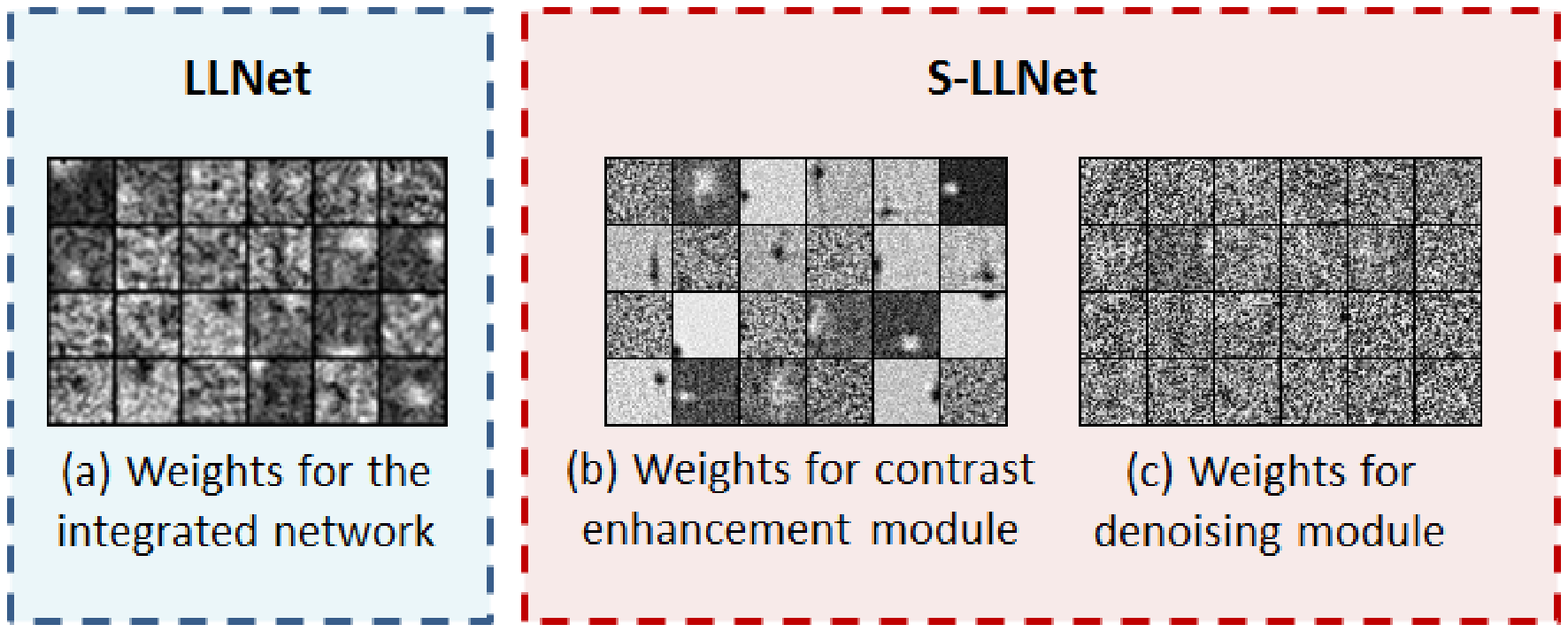}
\caption{Feature detectors can be visualized by plotting the weights connecting the input to the hidden units in the first layer. These weights are selected randomly.}
\vspace{-15pt}
\label{fig:weights}
\end{figure}



\textbf{Features of low-light images: }
To understand what features are learned by the model, weights linking the input to the first layer of the network can be visualized by plotting the values of the weight matrix as pixel intensity values (Fig.~\ref{fig:weights}). In a regular LLNet where both contrast enhancement and denoising are learned simultaneously, the weights contain blob-like structures with prominent coarse-looking textures. Decoupling the learning process (in the case of S-LLNet) allows us to acquire a better insight. Blob-like structures are learned when the model is trained for the task of contrast enhancement. The shape of the features suggest that contrast enhancement considered localized features into account; if a region is dark, then the model brightens it based on the context in the patch (i.e. whether the edge of an object is present or not). On the other hand, feature detectors for the denoising task appears noise-like, albeit in a finer-looking texture compared to the on coarser ones from the integrated network. These features shows that the denoising task is mostly performed in an overall manner. Note that while the visualizations presented by Burger et al. \cite{burger2012image} show prominent Gabor-like features at different orientations for the denoising task, these features are not apparent in the present study because training was done with varying noise levels. Specialization in different tasks enables S-LLNet to achieve a superior performance over LLNet at higher noise levels. 

\vspace{-8pt}
\section{Conclusions and future works}\label{sec:con}
A variant of the stacked sparse denoising autoencoder was trained to learn the brightening and denoising functions from various synthetic examples as filters which are then applied to enhance naturally low-light and degraded images. Results show that deep learning based approaches are suitable for such tasks for natural low-light images of varying degree of degradation. The proposed LLNet (and S-LLNet) compete favorably with currently used image enhancement methods such as histogram equalization, CLAHE, gamma adjustment, and hybrid methods such as applying HE first and subsequently using a state-of-the-art denoiser such as BM3D. While the performance of some of these methods remain competitive in some scenarios, LLNet (and S-LLNet) was able to adapt and perform consistently well across a variety of (lighting and noise) situations. This implies that deep autoencoders are effective tools to learn underlying signal characteristics and noise structures from low-light images without hand-crafting. Some envisaged improvements and future research directions are: (i) Training with Poisson noise and quantization artifacts to simulate a more realistic situation, (ii) include de-blurring capability explicitly to increase sharpness of image details; (iii) train models that are robust and adaptive to a combination of noise types with extension beyond low-light scenarios such as foggy and dusty scenes; (iv) perform subjective evaluation by a group of human users.


\bibliography{llnet_paper}
\bibliographystyle{icml2016}

\end{document}